# Beyond LLMs: Advancing the Landscape of Complex Reasoning

Jennifer Chu-Carroll†, Andrew Beck, Greg Burnham, David OS Melville, David Nachman, A. Erdem Özcan, David Ferrucci

Elemental Cognition

## Abstract

Since the advent of Large Language Models a few years ago, they have often been considered the de facto solution for many AI problems. However, in addition to the many deficiencies of LLMs that prevent them from broad industry adoption, such as reliability, cost, and speed, there is a whole class of common real world problems that Large Language Models perform poorly on, namely, constraint satisfaction and optimization problems. These problems are ubiquitous and current solutions are highly specialized and expensive to implement. At Elemental Cognition, we developed our EC AI platform which takes a neuro-symbolic approach to solving constraint satisfaction and optimization problems. The platform employs, at its core, a precise and high performance logical reasoning engine, and leverages LLMs for knowledge acquisition and user interaction. This platform supports developers in specifying application logic in natural and concise language while generating application user interfaces to interact with users effectively. We evaluated LLMs against systems built on the EC AI platform in three domains and found the EC AI systems to significantly outperform LLMs on constructing valid and optimal solutions, on validating proposed solutions, and on repairing invalid solutions.

## Introduction

In today's dynamic business landscape, the quest for competitive advantage compels enterprises to seek AI powered decision-making solutions to provide them with more advanced problem-solving capability and increased efficiency. As a result, many companies, veteran tech companies and start ups alike, are exploring ways to harness the power of generative AI in their own operations and product offerings.

Although Large Language Models (LLMs) have the remarkable capability of interacting fluently with humans – understanding user questions and statements and producing natural and convincing sounding responses, it is well known that LLMs suffer from the hallucination problem where unsubstantiated content is presented as facts (*Hughes Hallucination Evaluation Model*, n.d.). This is clearly a challenge both for businesses developing products involving generative AI and for businesses consuming those products. Despite passing an increasing number of tests which showcase their growing breadth and depth (OpenAI, 2023; Anthropic, 2023; Pichai & Hassabis, 2023), current LLMs lack the ability to perform sound logical inference and cannot reliably explain its underlying reasoning to support its own claims. As

†Corresponding author: jenniferc@ec.ai

Sam Altman recently said on the Hard Fork Podcast (*'Hard Fork': An Interview With Sam Altman*, 2023), "the main thing [LLMs] are bad at is reasoning, and a lot of the valuable human things require some degree of complex reasoning."

To better understand the truth of this statement and to quantify the reasoning capabilities of LLMs, which is pretty much the de facto representation of AI today, we set out to evaluate an LLM's performance on solving problems and validating their solutions on a selection of constraint satisfaction and optimization tasks that require heavy use of reasoning. We then compare the LLM's performance against those of systems developed on our EC AI platform.

Elemental Cognition's AI platform leverages the power of LLMs while enabling businesses to make reliable decisions in situations that require complex reasoning. We focus on complex reasoning because its application space is broad and of high value, including resource allocation and management, supply chain analysis and planning, architectural design and planning, financial portfolio construction, and so on. Our platform is built on a neuro-symbolic approach to AI where both neural and symbolic methods are used in different parts of the pipeline to marry the flexibility and fluency of language interaction of neural methods and the precision and reliability of reasoning of symbolic methods. We have developed real-world applications on our platform that interact with humans to solve their constraint satisfaction problems in complex travel planning and in undergraduate degree planning. The solutions our systems generate are guaranteed to maximally satisfy the user's requirements while staying within the bounds of the constraints of the problem domain.

To demonstrate the effectiveness of the EC AI platform as a problem-solving tool compared with that of LLMs, we conducted a comprehensive assessment of our platform against the current state-of-the-art LLM, namely GPT-4 (OpenAI, 2023). Our results demonstrate that our approach significantly outperforms GPT-4. Moreover, we argue that our hybrid platform is the more promising solution for solving complex reasoning problems over monolithic end-to-end large language models.

# LLMs and Reasoning

The advancement of generative AI and specifically LLMs have enabled language-driven AI-powered applications to grow by leaps and bounds in the past few years. LLMs have demonstrated their ability to write fluent essays, generate poetry, and even pass challenging standardized tests such as the US bar exam (OpenAI, 2023). Despite these impressive achievements, it is well known that LLMs suffer from the "hallucination" problem where the generated content may not be grounded in real world facts, leading to output containing false and/or self-contradictory information. This lack of reliability is a strong contributing factor for why LLMs are not widely adopted in industrial solutions today (Ilya Sutskever in Dwarkesh Podcast, 2023). Given this observation, how well can LLMs possibly perform on reasoning tasks? In other words, how reliably can LLMs ensure sound logic, make efficient inferences, guarantee optimal outcome, and produce explanations? It seems sensible to claim that if LLMs have good reasoning capabilities and a good internal model of factuality and truth, they would not be generating self-contradictory or ungrounded assertions in the first place. We set out to investigate how feasible it would be to develop solutions for problems that require reasoning using LLMs alone.

Figure 1: GPT-4 Response to Generating a 16x16 Sudoku Grid

First, we tried to solve some common puzzles involving constraint satisfaction with OpenAI's GPT-4 and found that the LLM is incapable of enforcing simple constraints on the solutions it generates. In Figures 1 and 2, we show GPT-4's responses to generating a 16x16 valid Sudoku grid and to solving the New York Times' Spelling Bee puzzle. For the Sudoku grid, while GPT-4 is able to explain the constraints correctly, it failed to generate a grid that abides by those constraints. Specifically, in Figure 1, column 0, which GPT-4 claims in its explanation contains all numbers exactly once, contains instead multiple instances of 0, 4, 8, and C and is missing all the other numbers. The Spelling Bee response in Figure 2 contains a word ("road") that does not contain the mandatory letter and incorrectly argues that the solution to this puzzle contains an infinite number of words.

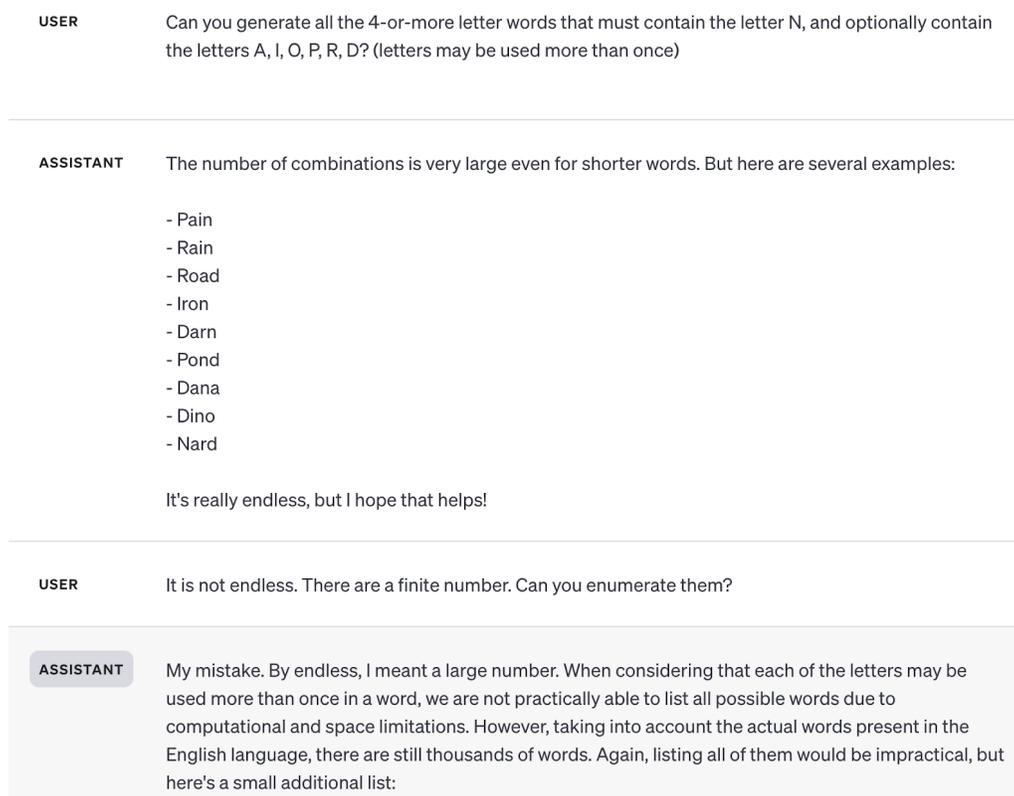

Figure 2: GPT-4 Response to a Spelling Bee Puzzle

These results clearly do not look promising. Not only was GPT-4 unable to produce a solution that satisfies all constraints of the puzzle, it failed to recognize that its solution violates some constraints while attempting to explain its validity; instead, in order to generate a coherent explanation, it presented data that is inconsistent with the solution it produced. Even though we have no reason to believe that LLMs today could attain reliable reasoning capabilities, we do believe that some amount of prompt engineering could improve its performance to a certain extent, as evidenced by the myriad of prompting techniques (Wei et al., 2022; Yao et al., 2023a; Wang et al., 2023; Yao et al., 2023b; Besta et al., 2023) and alternative approaches to leveraging LLMs (Tafjord et al., 2021) proposed to date to improve their reasoning capability. We will discuss later in the paper why we believe that this is not a sustainable long-term solution.

Given the broad array of practical applications that require constraint satisfaction and/or optimization, LLMs are clearly not going to solve these problems single-handedly. In the rest of this paper, we describe our hybrid approach that leverages LLMs and symbolic reasoning to produce reliable and provably correct solutions to this class of problems, and compare the performance of these two approaches on real world applications.

# A Neuro-Symbolic Approach to Solving Reasoning Problems

At Elemental Cognition, we take a neuro-symbolic approach to solving AI problems. Neuro-symbolic AI has been explored within the AI community in an attempt to overcome the deficiencies of neural methods and symbolic methods alone with various degrees of success so far, see e.g., (Olausson et al., 2023; Pan et al., 2023). Recently, Google DeepMind announced AlphaGeometry which takes a neuro-symbolic approach to solving geometry problems in the International Mathematics Olympiad scoring at a medal winner level (Trinh et al., 2024). We developed a platform which facilitates development of interactive systems that solve a host of constraint satisfaction and optimization problems. In this section, we give an overview of the EC AI platform and a few sample applications we have developed with this technology.

## Brief Introduction to the EC AI Platform

As depicted in Figure 3, at the heart of the EC AI platform is a general-purpose symbolic multi-strategy reasoning engine based on answer set programming built on top of Clasp (Gebser et al., 2012). The reasoning engine performs the key function of logical reasoning including causal, deductive, abductive and non-monotonic inference as well as multi-objective constraint optimization based on given rules and facts. It can determine whether or not the current state of the knowledge contains conflicting rules or facts and, if not, can compute an optimal solution given a series of optimization targets.

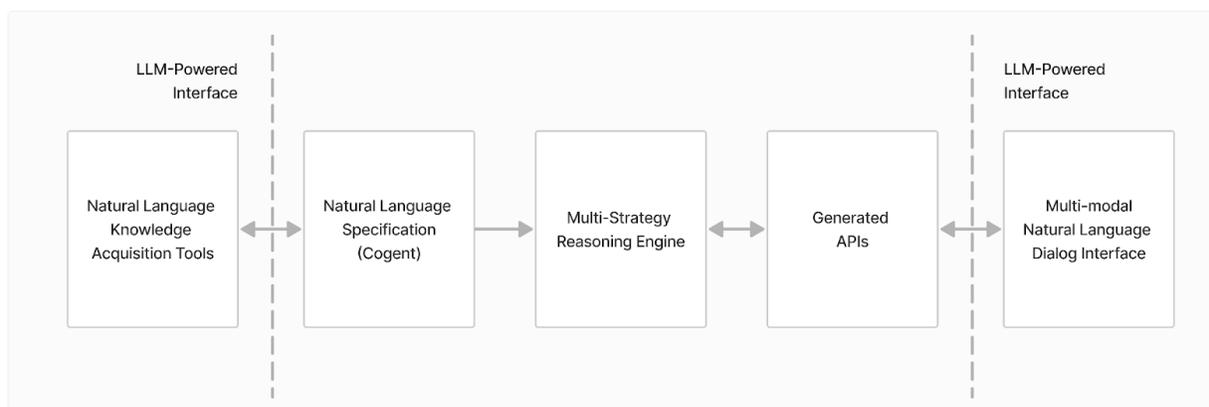

Figure 3: EC AI Platform Architecture, referred to as the "LLM Sandwich"

In order to ease the creation of applications that can reason and remove the need for expert programming skills, we developed a language, called **Cogent**, to serve as a bridge between the users and the reasoning engine. Cogent is a constrained subset of English with well defined semantics that allows users to capture optimization objectives as well as the rules and constraints governing the reasoning to be performed in unambiguous ways. It is processed by a compiler that can detect and rectify errors and report on logical inferences that need to be made to complete the understanding of an input specification, while allowing users to declare types and relationships explicitly to avoid possible incorrect inferences made by the compiler. Cogent is fundamentally a declarative language allowing the composition of complex specifications based on simpler specifications. Each sentence can be independently read and validated within the context of a specification irrespective of its length or complexity.

Figure 4: Sample Cogent Specification for Nurse Scheduling

Figure 4 shows a sample specification in Cogent for a nurse scheduling task. The specification contains a number of type and relationship definitions, which are in turn used to specify rules as well as constraints that must be satisfied for a solution to be valid. In addition, costs may be introduced to facilitate optimization among multiple valid solutions. The Cogent compiler compiles the model into a logical representation which can be directly consumed by our reasoning engine. Since this precise natural language specification is unambiguous and can be understood and processed by our formal reasoning process, our solution is guaranteed to produce correct, optimal answers even as task complexity increases.

Once a specification is defined, the EC AI platform automatically generates APIs that allow an external application to communicate directly with the reasoning engine, which is primed with the specified knowledge, to assert or remove facts, query the current state of the knowledge base, or request the reasoning engine to perform an optimization. These APIs can be used for connecting the application with real time data as well as for building structured user interfaces to allow users to interact with the application.

In addition to this core reasoning prompted with natural language and structured interaction capabilities, the EC AI platform integrates with LLM-powered interfaces to facilitate the knowledge acquisition and user interactions through fluid natural language interactions (see Figure 3). On the knowledge acquisition front, authors are provided with a suite of tools that enable a systematic and iterative workflow. The workflow starts with an interactive LLM-powered assistant, which aids application developers in translating their instructions from unconstrained natural language into precise Cogent specifications. This process is further augmented by the utilization of symbolic error detection, rectification, and diagnostic tools. On the user interaction front, a natural language dialog interface is automatically generated based on the knowledge specified by the user as well as the formal understanding of that specification by the Cogent compiler. This dialog interface, combined with structured graphical interfaces that can be built using the auto-generated APIs, enable the creation of rich and reliable multimodal interfaces where user interactions are not limited to user inputs that are necessarily processed with statistical tools as in the case of LLMs. We refer to our EC AI architecture where a core symbolic reasoning engine leverages LLM-enabled interfaces for knowledge acquisition and user interaction as the "**LLM Sandwich**".

## Sample Complex Reasoning Problems

Many real world applications require constraint satisfaction or optimization as part of their solutions. Today these applications are typically implemented using very specialized optimization engines by expert developers such as Gurobi (Gurobi Optimization, LLC, n.d.) and Google's OR tools (Google, n.d.) or in some cases, the tasks are still carried out manually. We have identified several application areas where the EC AI platform makes a significant impact. We discuss some of these application areas along with systems we have developed or are developing.

### Workforce Planning

Workforce planning covers problems where a limited set of resources need to be assigned to perform a set of tasks over a period of time to satisfy an organizational need. The tasks often come with requirements which the individual resources may or may not meet. Examples of workforce planning problems include shift scheduling in hospitals and restaurants as well as project resource planning in typical project-based organizations.

EC is currently developing an intelligent application for project planning for an organization. While we do not yet have an end-to-end application, we have a prototype that assigns employees to projects based on specifications from our customer.

### Complex Travel Planning

We refer to complex travel planning as the building of itineraries that are highly constrained in order to coordinate transportation, accommodation, and date requirements, to coordinate multi-party travel, or to conform to rules required to qualify for special fares. OneWorld airline alliance's Round the World (RTW) program is one such complex travel planning example. The program specifies a host of constraints that a valid itinerary must satisfy, including the minimum/maximum number of cities visited in total and in each geographic region, city ordering constraints based on geographic regions, very limited ticketing classes on each flight, and so on. According to statistics from OneWorld, as of 2020, the vast majority of these very complex itineraries are still being booked through human travel agents, since existing automated tools are often unable to guide the users to a satisfactory outcome given the complexity of the rules.

EC has developed and deployed an interactive, multi-modal conversational system for booking RTW tickets.[1] The rules and constraints of the program are codified and provided to the reasoning engine as domain knowledge along with background knowledge about geographical locations and air travel. The system attempts to optimize the travel route given program constraints and user requests. If no valid itinerary can be constructed, the reasoning engine pinpoints the constraint(s) being violated, allowing the system to guide the user to rectify the problem.

## Degree Planning

The degree planning application area covers systems that play the role typically carried out by an academic advisor today in higher education institutions. A degree planning assistant can help students at these institutions explore their major/minor/course options and construct or validate a course plan to complete their degree(s) by a target date. The plan considers a student's current progress against degree conferral requirements, course pre-/co-requisites, and planned future course offerings. This role is typically carried out by academic advisors today which is expensive, inefficient, and most of all, error prone.

EC has developed and deployed an interactive system for degree planning at a university which was reported to be among the global top 50 according to US News.[2] The university's degree conferral requirements as well as departmental course offerings and dependencies are codified and made available to the reasoning engine. Student transcripts are loaded upon log in and used as the basis for planning. Students can ask the system to help build, validate, or modify a plan. They can also ask the system for course recommendations or suggestions of suitable minors given a plan.

# Constructing and Validating Plans with EC AI vs LLMs

We set out to evaluate the effectiveness of the EC AI platform for solving complex reasoning and optimization problems and for validating its solutions as compared with OpenAI's GPT-4. We recognize that the EC AI platform, with its use of a logical reasoning engine, is designed to focus on solving the class of complex reasoning problems discussed here while LLMs are general purpose models that cover far more breadth than the EC AI systems. We focus in this paper on evaluating EC AI systems against GPT-4's performance today, recognizing that LLMs are continuously improving but that there are fundamental limitations to the monolithic end-to-end nature of LLMs.

## Evaluation Criteria

We argue that in order for a system to be considered adept at solving constraint satisfaction and optimization problems and to be useful in assisting humans, it must have the following capabilities:

---

[1] https://rtw-va.oneworld.com/
[2] https://www.usnews.com/education/best-global-universities/rankings. Due to contractual obligations, we are unable to divulge the name of the university at this time and cannot provide a URL to the deployed system.

1. **Optimal plan construction**: given a set of input, the system can construct an optimal solution given the domain and user-provided constraints and should probe for missing information necessary to compute a solution.
2. **Plan validation**: given a proposed solution, the system can determine if the solution is valid according to the given constraints.
3. **Plan repair**: when told that a solution is invalid, the system can identify the violated constraint(s) and modify the solution to become valid.
4. **Answer explanation**: the system can introspect and provide a concise summary of its reasoning process to support any answer it gives to the user.
5. **Answer reliability**: given the same input twice, the system produces the same logical response.

We focus our evaluation on the first three requirements above. We did not systematically evaluate answer reliability because anecdotally we observe that the EC AI systems always produce the same logical output for the same input, whereas it is well known that GPT-4 is highly non-deterministic (even with a fixed seed and temperature set to 0). We informally assessed the systems' ability to explain its reasoning and found GPT-4 to suffer from the same hallucination problems as shown in the Sudoku explanation in Figure 1, while the reasoning engine in the EC AI systems is able to produce a full reasoning graph for valid solutions and pinpoint the exact constraints that are violated for invalid solutions.

Given our experience in building applications using the EC AI Platform as well as in experimentation with GPT-4 on this class of problems, we expect the EC AI systems to significantly outperform GPT-4. This is because EC's approach decomposes the end-to-end problem in an application into multiple stages, leveraging best-in-class technologies to solve the problem presented at each stage. A symbolic reasoner is known to be precise and robust and is built for solving constraint satisfaction and optimization problems. On the other hand, to the extent that LLMs can produce correct solutions, they will be doing so using a monolithic blackbox that was trained to perform optimally on distantly related tasks.

# Experimental Setup

## Systems and Baselines

We set out to compare three systems, one in each of the three domains above, against an LLM performing the same tasks. The three systems used in the evaluation are:

1. **Projecto** for workforce planning: a prototype system developed for an EC customer for project planning.
2. **RTW** for complex travel planning: EC's deployed system for booking Round the World travel on the OneWorld website. To simplify the evaluation, we only considered the city ordering portion of the planning task to eliminate the need to interactively provide real time flight data information to the LLM.
3. **Classpath** for degree planning: EC's deployed system at a university to help students create a plan for earning a Bachelor's degree. Again, due to the complexity of the task, the evaluation is simplified to consider plans for earning a Bachelor's of Business degree in 8 majors only.

For all three systems, a Cogent document is used as the specification of the problem domain, and each test instance is mapped to a user request in English for the system to carry out. After each user interaction, the reasoning engine is queried for an optimal solution if it exists, or for the rule violation or missing information that prevents a solution from being found.

For comparison, we configured 3 GPT-4 instances such that each system prompt includes specifications of the business rules that have been used to produce a reliable high-performing system for the application, along with instructions for the task that GPT-4 is asked to carry out using those business rules.

To evaluate the three pairs of systems for their performance using the criteria identified above, we carried out two experiments: **plan construction** and **plan repair**. For **plan construction**, both the EC AI systems and GPT-4 are given a set of user requirements and optimality criteria and are asked to produce the best plan. System output is judged based on whether a plan is optimal, valid, or invalid. For **plan repair**, both systems are given invalid plans and asked to consult the business rules to determine its validity. For plans deemed invalid, the systems are asked to make minimal modification to the plan to make it valid.

## Evaluation Data

To the extent possible, we used real user data collected from our deployed systems for our evaluation. We also varied the complexity in the examples in the test set to evaluate how system performance varied based on task complexity. We describe the source and characteristics of our test set for plan construction for each system below:

1. **Projecto**: A dataset consisting of 20 projects and 10 employees was used as the basis for constructing the test set. Tasks of varying complexity were automatically created by randomly choosing m projects and n employees for some pre-selected m and n. Test instances with no possible solution were filtered out, resulting in a test set of 87 instances.
2. **RTW**: We extracted 137 itineraries from the logs of our deployed system. We converted these itineraries into user requests by extracting the home city and randomizing the rest of the cities in the itinerary. The task complexity naturally varies within the examples we extracted in terms of number of cities visited, the geographic regions the cities are located in, and connectivity between the cities.
3. **Classpath**: A valid transcript for a Bachelor's degree in each of the 8 majors was used as the seed. Tasks of varying complexity were automatically created by removing the last n semesters in the transcript and asking the system to create a plan for those n semesters sequentially for n = 1 to 5.[3] This resulted in 40 test instances.

For plan repair, the evaluation set consists of the invalid plans produced by GPT-4 during plan construction. The test sets for this task are smaller across the board, with 47 examples for Projecto, 20 examples for RTW, and 36 examples for Classpath.

---

[3] The university we worked with is on a 3-year system.

# Evaluation Results

For the plan construction evaluation, plans produced by the EC AI systems and by GPT-4 are judged on both validity (whether or not a plan conforms to all rules given) and optimality (whether or not a better plan exists, based on optimality criteria given, such as the time needed to complete all projects). Table 1 shows GPT-4's plan construction performance on producing valid and optimal solutions. The first row shows the performance on the most complex examples in our test sets, namely having more projects and employees to assign in Projecto, having more cities in the itinerary in RTW, and having more remaining semesters to schedule in Classpath. Each successive row includes additional simpler examples until the last row which uses all test instances.

For the EC AI systems, our knowledge capture tools allow developers to accurately model application rules and constraints in the Cogent language. Furthermore, our multimodal interactive systems engage with the users and leverage LLMs to interpret user requests as calls to APIs generated from the model. This interpretation layer works with high reliability for straightforward user requests. With those capabilities, coupled with our precise and deterministic logic-based reasoning engine, the EC AI systems produced valid and optimal solutions 100% of the time on all three tasks.

Table 1: GPT-4 performance on constructing plans, evaluated based on plan validity and plan optimality

|  | **Projecto** (workforce planning) | | **RTW** (travel planning) | | **Classpath** (degree planning) | |
| --- | --- | --- | --- | --- | --- | --- |
|  | **Valid** | **Optimal** | **Valid** | **Optimal** | **Valid** | **Optimal** |
| **20% most complex** | 0% | 0% | 67% | 11% | 0% | 0% |
| **40% most complex** | 0% | 0% | 62% | 21% | 6% | 6% |
| **60% most complex** | 4% | 0% | 62% | 29% | 8% | 8% |
| **80% most complex** | 11% | 0% | 70% | 35% | 6% | 6% |
| **All test instances** | 46% | 0% | 77% | 51% | 10% | 10% |

For the plan repair experiment, the test cases are all invalid plans, and the systems are evaluated based on whether it correctly classified a plan as invalid, and of those classified as invalid, whether it was able to modify the plan to become valid. Table 2 shows GPT-4's plan validation and plan repair performance on all three applications. As with the earlier results, the systems built with the EC AI Platform achieve 100% on all three tasks.

Table 2: GPT-4 performance on plan validation and plan repair

|  | **Projecto** (workforce planning) | **RTW** (travel planning) | **Classpath** (degree planning) |
|---|---|---|---|
| **Plan Validation** | 2% | 75% | 42% |
| **Plan Repair** | 0% | 47% | 7% |

## Results Analysis

The plan construction results in Table 1 show that GPT-4 is incapable of producing valid solutions for the most complex scenarios in our workforce planning and degree planning applications. In fact, for workforce planning, GPT-4 only approached 50% valid when the simplest of all scenarios are included (the simplest test cases plan only for 2 projects), and all the solutions it generated were suboptimal, meaning that the total time to project completion was longer than the optimal case. GPT-4 performance on degree planning shows similarly poor performance, except that GPT-4 is only able to produce valid plans 10% of the time when all test cases are considered. On the other hand, when GPT-4 was able to put together a valid plan in this application, it was always optimal (scheduled students to finish in as few semesters and left as many electives open as possible). The outlier application here is clearly travel planning, where GPT-4 is able to produce valid itineraries for the majority of test cases; however, when optimality (travel distance) is considered, its performance drops quite significantly. We believe that the reason why GPT-4 is able to construct valid itineraries quite easily is because, for most itineraries, simply visiting all cities in one continent before moving to the next constitutes a valid itinerary.[4][5]

For plan repair (Table 2), GPT-4 identified less than half of the invalid plans as invalid in degree planning, and only was able to repair a very small number of them. It performed abysmally in workforce planning, practically failing on both tasks. GPT-4 again performed better on travel planning. It was able to correctly identify 75% of all test examples as invalid (although the explanations it gave for violated constraints were mostly incorrect). It was also able to repair roughly half of those recognized invalid solutions, which we believe is due to the same reasons that it was able to generate correct solutions off the bat more frequently.

---

[4] The scheduling complexity in this evaluation comes when the user wants to visit too many cities in one continent or wants to visit the cities in a certain order. The real complexity in this task occurs later on in the booking process when selecting dates and flights, which we did not tackle in this evaluation due to the complexity in providing real time flight data to the LLM.

[5] In fact, in a second GPT-4 evaluation, we replaced the RTW fare rules in the system prompt by the instruction to group all cities in the same continent together and with those instructions, GPT-4 generated valid plans for 84% of all test cases, versus 77% with RTW fare rules.

# Discussions

EC AI's remarkable performance can be attributed to its distinctive approach of separating the mechanisms for knowledge capture and reasoning execution. This clear distinction sets it apart as an AI platform, in contrast with LLMs that integrate them into an end-to-end framework. Precise knowledge capture is fundamental to building a system that reliably produces accurate outcomes and is readily introspectable. EC AI facilitates this through the utilization of the Cogent language, which combines well-defined semantics and a deterministic process which converts specifications in Cogent into logical forms that can be consumed by the reasoning engine.

Our experiments comparing systems built with EC AI against GPT-4 have demonstrated that our systems significantly outperform GPT-4 in plan construction, plan validation, and plan repair. The key factors that contribute to EC AI's superior performance is summarized in the following.

First, Cogent, the language that holds the EC AI platform together, is a declarative language designed for systematically composing statements to capture complex knowledge with simplicity. Each sentence can be independently read and validated, irrespective of the complexity or length of the captured knowledge. Furthermore, the language is meticulously designed to eliminate ambiguity, a common pitfall of fully expressive natural language. This concept holds significant power, as it inherently facilitates scalability in addressing complex reasoning scenarios.

Second, knowledge capture is supported by a suite of tools that enable a systematic and iterative workflow. This workflow starts with an LLM-powered assistant, aiding application developers in translating their instructions from unconstrained natural language into precise language. Given the rigorous semantics of Cogent, a compiler processes this language, further helping developers by identifying and rectifying inconsistencies and errors.

Finally, the execution of these precise specifications by a symbolic reasoning engine ensures the deterministic generation of output and provides explanations for results, thereby establishing a clear link between the output and the knowledge captured in the specifications through first principles reasoning. This framework enables diagnostic and validation tools to further aid users in comprehending system behavior and engaging in iterative refinement of the specifications, enabling the creation of applications that fully align with expectations.

The performance of EC AI systems does come with a higher development cost than prompting an LLM to produce a solution. We reduce the development effort by providing tooling to help with application building. On the knowledge capture front, we have created a developer-oriented author assistant to help convert English language descriptions into valid Cogent specifications with links to relevant sections of the Cogent documentation. On the user interaction front, we have developed the capability of configuring a zero-shot dialogue experience to interact with application users to query or modify the current state of the application solution. We continue to invest in improving our tooling and user experience to make the EC AI platform more accessible to developers and non-developers alike.

Compared with EC AI's clear specifications and grounded execution, the opaque and statistical model of LLMs makes a systematic engineering process unattainable. The inherent lack of explainability in LLMs poses significant challenges in establishing a clear link between the output and the knowledge encapsulated in the prompt. Moreover, the presence of nondeterminism in the output further complicates analysis as running an LLM with the same input multiple times can yield entirely different results.

Additionally, the token-by-token processing approach employed by LLMs makes it impossible to systematically compose more complex prompts from simpler ones. This limitation often results in situations where improving knowledge capture to align an application's behavior with expectations is impractical. Instead, users resort to empirical methods such as few-shot prompting, chain of thought prompting, or prompt chaining, etc. to enhance application accuracy (*Prompting Techniques*, 2024). Furthermore, because LLMs have no formal representation to which natural language input is mapped and no mechanism that systematically converts an input to a deterministic output, there is no "recipe" for when and how to adopt the various prompt improvement techniques. This trial-and-error approach, lacking guaranteed incremental progress toward the expected outcome, deviates significantly from the discipline of engineering, contrary to the implication of the term "prompt-engineering."

Overall, while we recognize that LLMs are continuously improving and there is active work being carried out to better leverage LLMs for solving reasoning problems, we believe that there are fundamental limitations to the current monolithic end-to-end purely statistical approach to training LLMs on human language. Instead, new architectures, such as the approach taken in the EC AI platform, will be required to overcome the limitations.

# Conclusions

In this paper, we demonstrated that LLMs, despite their remarkable abilities at generating coherent prose in natural language, are severely lacking in producing satisfactory results for solving constraint satisfaction and optimization problems. These problems are ubiquitous in real world applications but developer-friendly and effective platforms for building solutions to these problems are still lacking today.

We described the EC AI platform which was developed to support development of business applications that involve constraint satisfaction and optimization in their solutions. The platform's architecture, referred to as the LLM Sandwich, employs a precise symbolic reasoning engine at its core. It includes LLM-powered knowledge capture tools to facilitate specification of rules and constraints for an application in our constrained English language, Cogent. Cogent specifications are compiled into logical forms that our reasoning engine can directly ingest and use for problem solving at run time. The platform also utilizes LLMs for support in developing an interactive user interface to the application through an API automatically generated from the Cogent specifications. The LLMs are leveraged in these developer tools to support a natural and effective experience in the development of the applications.

We evaluated GPT-4 against three applications built with the EC AI platform for workforce planning, travel planning, and degree planning. We evaluated the systems' capabilities to construct a plan given a set of inputs and optimization criteria, as well as their abilities to validate a plan against a set of constraints and to rectify the plan if it is found to be invalid. Our results show that across all three applications, the EC AI systems, with LLM-support for user interaction, were able to achieve 100% performance in all dimensions, significantly outperforming GPT-4, which was able to produce some valid solutions, the vast majority of which were suboptimal, and performed poorly in recognizing or rectifying invalid solutions.

# References

Anthropic. (2023, July 11). *Claude 2*. Anthropic. Retrieved January 29, 2024, from

https://www.anthropic.com/news/claude-2

Besta, M., Blach, N., Kubicek, A., Gerstenberger, R., Gianinazzi, L., Lehmann, T., Podstawski, M.,

Niewiadomski, H., Nyczyk, P., & Hoefler, T. (2023). *Graph of Thoughts: Solving Elaborate*

*Problems with Large Language Models*. https://arxiv.org/pdf/2308.09687.pdf

Dwarkesh Podcast. (2023, March 27). *Ilya Sutskever (OpenAI Chief Scientist) - Building AGI, Alignment,*

*Spies, Microsoft, & Enlightenment*. YouTube. Retrieved February 1, 2024, from

https://www.youtube.com/watch?v=Yf1o0TQzry8

Gebser, M., Kaufmann, B., & Schaub, T. (2012). Conflict-Driven Answer Set Solving: From Theory to

Practice. *Artificial Intelligence*, *187*, 52-89.

https://www.cs.uni-potsdam.de/wv/publications/DBLP_journals/ai/GebserKS12.pdf

Google. (n.d.). *Google OR-Tools*. Google for Developers. Retrieved February 1, 2024, from

https://developers.google.com/optimization

Gurobi Optimization, LLC. (n.d.). Gurobi Optimizer Reference Manual. Retrieved February 1, 2024,

from https://www.gurobi.com

*'Hard Fork': An Interview With Sam Altman*. (2023, November 24). YouTube. Retrieved January 26,

2024, from https://www.youtube.com/watch?v=wBX4xeefPiA

*Hughes Hallucination Evaluation Model*. (n.d.). Hugging Face. Retrieved January 26, 2024, from

https://huggingface.co/spaces/vectara/leaderboard

Olausson, T., Gu, A., Lipkin, B., Zhang, C., Solar-Lezama, A., Tenenbaum, J., & Levy, R. (2023). LINC: A Neurosymbolic Approach for Logical Reasoning by Combining Language Models with First-Order Logic Provers. *Proceedings of the Conference on Empirical Methods in Natural Language Processing*. https://aclanthology.org/2023.emnlp-main.313/

OpenAI. (2023). GPT-4 Technical Report. https://cdn.openai.com/papers/gpt-4.pdf

Pan, L., Albalak, A., Wang, X., & Wang, W. Y. (2023). Logic-LM: Empowering Large Language Models with Symbolic Solvers for Faithful Logical Reasoning. *Findings of the Conference on Empirical Methods in Natural Language Processing*. https://arxiv.org/abs/2305.12295

Pichai, S., & Hassabis, D. (2023, December 6). *Introducing Gemini: Google's most capable AI model yet*. The Keyword. Retrieved January 29, 2024, from https://blog.google/technology/ai/google-gemini-ai/

*Prompting Techniques*. (2024, January 11). Prompt Engineering Guide. Retrieved January 26, 2024, from https://www.promptingguide.ai/techniques

Tafjord, O., Mishra, B. D., & Clark, P. (2021). ProofWriter: Generating Implications, Proofs, and Abductive Statements over Natural Language. *Findings of the Association for Computational Linguistics*. https://aclanthology.org/2021.findings-acl.317.pdf

Trinh, T. H., Wu, Y., Le, Q. V., He, H., & Luong, T. (2024, January 18). Solving olympiad geometry without human demonstrations. *Nature*, *625*, 476-482. https://doi.org/10.1038/s41586-023-06747-5

*2022-2023 Best Universities in the World*. (n.d.). USNews.com. Retrieved January 29, 2024, from https://www.usnews.com/education/best-global-universities/rankings

Wang, X., Wei, J., Schuurmans, D., Le, Q., Chi, E. H., Narang, S., Chowdhery, A., & Zhou, D. (2023). Self-Consistency Improves Chain of Thought Reasoning in Language Models. *Proceedings of the International Conference on Learning Representations*. https://arxiv.org/pdf/2203.11171.pdf

Wei, J., Wang, X., Schuurmans, D., Bosma, M., Ichter, B., Xia, F., Chi, E. H., Le, Q. V., & Zhou, D. (2022). Chain-of-Thought Prompting Elicits Reasoning in Large Language Models. *Proceedings *of the 36th Conference on Neural Information Processing Systems*.

https://arxiv.org/pdf/2201.11903.pdf

Yao, S., Yu, D., Zhao, J., Shafran, I., Griffiths, T. L., Cao, Y., & Narasimhan, K. (2023a). Tree of Thoughts: Deliberate Problem Solving with Large Language Models. *Proceedings of the 37th Conference of Neural Information Processing Systems*. https://arxiv.org/pdf/2305.10601.pdf

Yao, S., Zhao, J., Yu, D., Du, N., Shafran, I., Narasimhan, K., & Cao, Y. (2023b). ReAct: Synergizing Reasoning and Acting in Language Models. *Proceedings of the International Conference on Learning Representations*. https://arxiv.org/pdf/2210.03629.pdf